\documentclass[conference]{IEEEtran}
\IEEEoverridecommandlockouts
\usepackage{cite}
\usepackage{amsmath,amssymb,amsfonts}
\usepackage{algorithmic}
\usepackage{graphicx}
\usepackage{textcomp}
\usepackage{xcolor}
\usepackage{listings}
\def\BibTeX{{\rm B\kern-.05em{\sc i\kern-.025em b}\kern-.08em
    T\kern-.1667em\lower.7ex\hbox{E}\kern-.125emX}}

\usepackage{subcaption}   

\begin{document}

\bibliographystyle{IEEEtran}

\title{Efficient Cloud Pipelines for Neural Radiance Fields\\

}

\author{\IEEEauthorblockN{Derek Jacoby}
\IEEEauthorblockA{\textit{Dept of Computer Science} \\
\textit{Univ of Victoria}\\
Victoria, Canada \\
0000-0002-1552-7484}
\and
\IEEEauthorblockN{Donglin Xu}
\IEEEauthorblockA{\textit{Dept of Computer Science} \\
\textit{Northeastern University}\\
Vancouver, Canada \\
}
\and
\IEEEauthorblockN{Weder Ribas}
\IEEEauthorblockA{\textit{Dept of Computer Science} \\
\textit{Northeastern University}\\
Vancouver, Canada \\
}
\and
\IEEEauthorblockN{Minyi Xu}
\IEEEauthorblockA{\textit{Dept of Computer Science} \\
\textit{Northeastern University}\\
Vancouver, Canada \\
}
\and
\IEEEauthorblockN{Ting Liu}
\IEEEauthorblockA{\textit{dept. of Computer Science} \\
\textit{Univ. of Victoria}\\
Victoria, Canada \\
}
\and
\IEEEauthorblockN{Vishwanath Jeyaraman}
\IEEEauthorblockA{\textit{Dept of Computer Science} \\
\textit{Northeastern University}\\
Vancouver, Canada \\
}
\and
\IEEEauthorblockN{Mengdi Wei}
\IEEEauthorblockA{\textit{Dept of Computer Science} \\
\textit{Northeastern University}\\
Vancouver, Canada \\
}
\and
\IEEEauthorblockN{Emma De Blois}
\IEEEauthorblockA{\textit{dept. of Computer Science} \\
\textit{Univ. of Victoria}\\
Victoria, Canada \\
}
\and
\IEEEauthorblockN{Yvonne Coady}
\IEEEauthorblockA{\textit{dept. of Computer Science} \\
\textit{Univ. of Victoria}\\
Victoria, Canada \\
ycoady@uvic.ca}
}

\maketitle

\begin{abstract}
Since their introduction in 2020, Neural Radiance Fields (NeRFs) have taken the computer vision community by storm. They provide a multi-view representation of a scene or object that is ideal for eXtended Reality (XR) applications and for creative endeavors such as virtual production, as well as change detection operations in geospatial analytics. The computational cost of these generative AI models is quite high, however, and the construction of cloud pipelines to generate NeRFs is neccesary to realize their potential in client applications. In this paper, we present pipelines on a high performance academic computing cluster and compare it with a pipeline implemented on Microsoft Azure. Along the way, we describe some uses of NeRFs in enabling novel user interaction scenarios.

\end{abstract}

\begin{IEEEkeywords}
Neural Radiance Fields, Cloud Computing, Azure, Containerization
\end{IEEEkeywords}

\section{Introduction}
Neural Radiance fields (NeRFs) are a trained multi-view representation of a scene. First introduced by Mildenhall et al \cite{mildenhall_nerf_2020}, NeRFs have more recently received much attention in the Computer Vision community \cite{gao_nerf_2022}. In this paper, we document the process of collecting aerial imagery data and processing it into point clouds and NeRFs. We do so on two platforms, initially on Compute Canada (also known as the Digital Research Alliance of Canada, DRAC) and then later on Microsoft Azure to overcome some of the limitations of an academic High Performance Computing (HPC) environment.

Commercial cloud platforms, such as Azure, offer several advantages over academic HPC clusters. They provide on-demand scalability, cost-effectiveness, and enhanced accessibility globally. \cite{zhao_enabling_2015}  Also offered is a wide variety of complementary services including data storage, networking, and machine learning tools, making them valuable for developing and deploying NeRF applications. To address these limitations, this research paper aims to showcase an alternative approach by migrating the pipeline to a commercial cloud computing platform, specifically Microsoft Azure. When compared with commercial cloud platforms counterparts like Amazon Web Services (AWS), Microsoft Azure is a more appropriate option considering pricing and availability according to \cite{al2019investigation}. \cite{ding2013hpc} experimented deploying a large scale computation intensive application.

The infrastructure presented here specifically targets large-scale datasets, such as representations of urban geometry acquired by remote sensing, and processing pipelines to make use of that data. NeRFs offer a perceptually-based storage mechanism that is supportive of photorealistic renderings in a highly multi-level geospatial dataset \cite{xiangli_bungeenerf_2022}, used for efficient representations of large urban scenes \cite{turki_mega-nerf_2022}. This is a very different set of affordances from traditional point cloud data, although attempts have been made to merge the advantages of both methods of representing the underlying scene \cite{rematas_urban_2022}. Although our pipeline work is represented in terms of a geospatial analytics application, the technology lends itself to powerful experiences in virtual production through NeRF intergrations in platforms such as Unity and Unreal Engine. Here is an end-user focused description of the technology and the user experience it enables.

Over two hours north of British Columbia’s capital city, bordering Cameron Lake, a visitor explores the ancient fir trees of Cathedral Grove that shelter a lush ecosystem of trails, brush, and waterway. Orange-crowned warblers sing from their perches, and sunlight breaks through the tree canopy in hazy, bright beams. At the end of one trail is a wide Douglas Fir, the widest in the park. The visitor embraces its trunk and its well-worn bark, then examines a fallen totem and learns about its significance to the local Nuu-chah-nulth (colloquially, Nootka) First Nations. Foot traffic in the popular park has an undeniable impact on the ecosystem, habitats are often trampled by visitors who venture off of paths. This visitor will have no negative impact on the land; they have come to it from within four walls, with their VR headset on. \\

They are exploring a multi-view representation of Cathedral Grove, generated using Neural Radiance field (NeRF) technologies. This powerful 3D representation, built using 2D images and generative AI, can be navigated freely. The camera is not limited to a set path but rather allows for improvisation and exploration within a setting. When these technologies are used to create stages for applied and/or site-specific theatre, the opportunities to create engaging, interactive, and immersive stories are endless. 

Although the example of Cathedral Grove is hypothetical, in this paper, we will describe our process for creating scene NeRFs from drone imagery, including an open-source cloud-based pipeline for automating the creation of NeRFs. We will explore the use of photogrammetry and NeRFs for the creation of 3D object assets and identify the challenges and solutions in building a containerized Nerfstudio \cite{tancik_block-nerf_2022} pipeline in a High Performance Computing (HPC) environment and on Microsoft Azure. With the participation of a museum, we have 3D scanned assets of historical objects to incorporate into digital stories in these captured scenes. In a brief discussion, we will describe current work in establishing a budget virtual production studio and using it to enable digital storytelling.

\section{Related work}

We begin this survey with some background in photogrammetry and point clouds, a brief discussion of quality metrics, and some additional NeRF rendering work.

{\em Photogrammetry and point clouds}:  As is common for point cloud projects in this space, we make heavy use of the COLMAP project \cite{schoenberger2016sfm}. Our initial work on quality metrics shows favourable comparison with commercial tools, although a recent comparative review of structure-from-motion tools offers a more nuanced comparison \cite{keyvanfar_performance_2022}. In our own quality assessment work we rely on recent no-reference methods and present only our preliminary results \cite{zhang_no-reference_2022} and compare our results with those of the SensatUrban project \cite{hu2020towards}.

As traditional NeRF rendering continues to improve on commodity PCs, mobile devices are coming into reach. Chen et al. introduced a new representation of NeRF - MobileNeRF \cite{chen_mobilenerf_2022} based on textured polygons for efficient synthesis of 3D scene images using standard rendering pipelines. Interactive frame rates were achieved on a variety of platforms through parallelism provided by traditional polygon rasterization.  Tradeoffs include challenges involving scenes with specular surfaces and/or sparse views, along with no semi-transparencies and texture resolutions.  Cao et al. \cite{cao2022real} provides a network that runs in real-time on mobile devices for neural rendering. It achieved similar image quality as NeRFs and better quality than MobileNeRF on a real-world dataset.

{\em Cloud Rendering and Client Rendering}: Client rendering of NeRFs often requires hardware that is costly and inconvenient. Park et al. propose a hybrid method \cite{park_instantxr_2022} that fuses traditional 3D asset rendering with NeRF technology to create and display XR environments in real-time using photographs, without the need for a modeling process, to expedite the XR environment creation process.  Given that hardware requirements for NeRFs are a limitation, cloud-based rendering services, which allow users to access high-performance rendering without needing to invest in expensive hardware has potential. A recent study on cloud and lightweight rendering \cite{zhang_cloud--end_2020} conducted by Zhang et al. showed that their cloud-to-end rendering and storage system provided satisfying rendering while keeping computation costs under control. 

\section{Experimental Setup and Pipelines}
\begin{figure}[h]

    \centering
    \begin{subfigure}{0.3\textwidth}
        \centering
        \includegraphics[width=\textwidth]{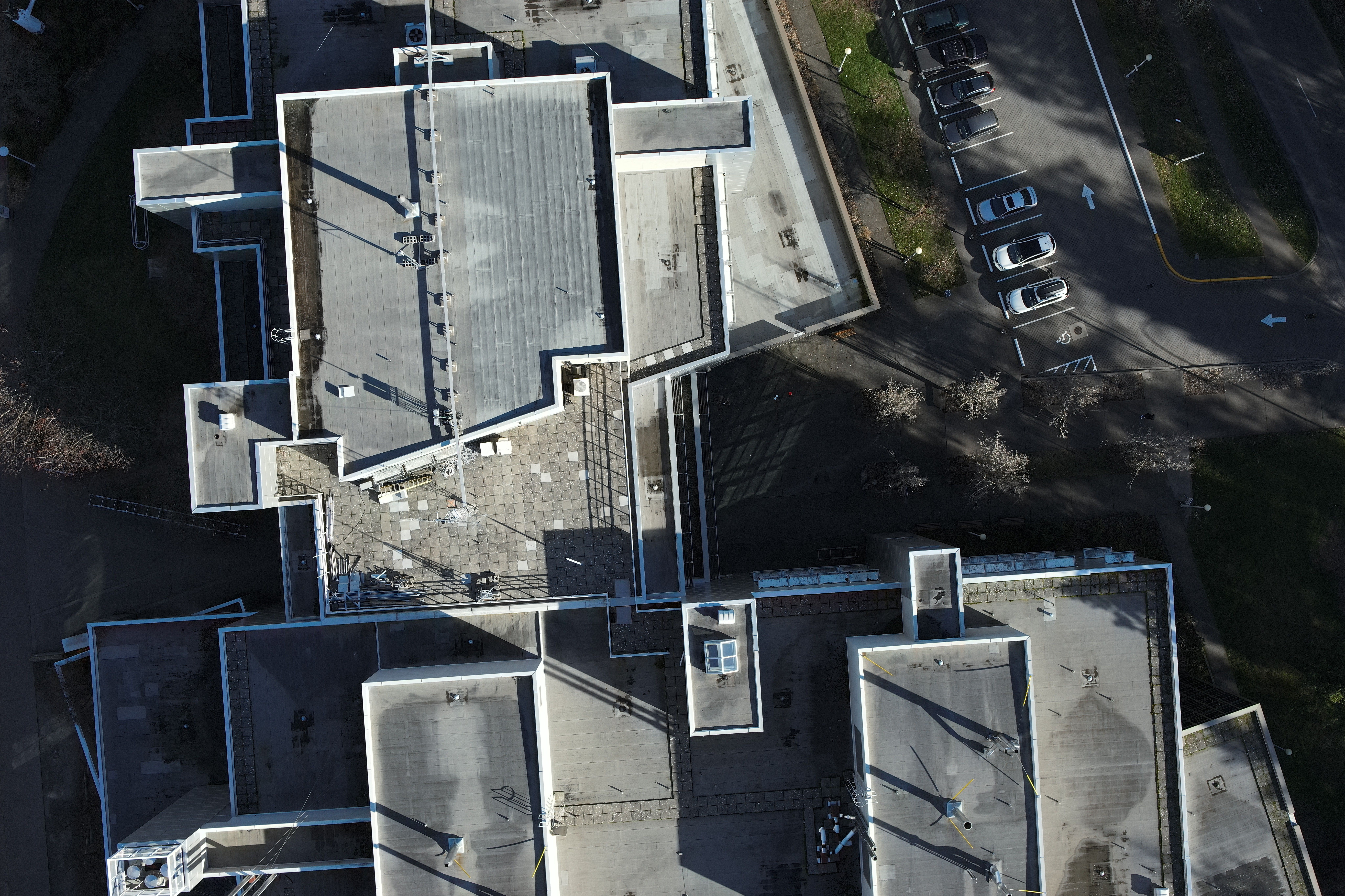}
        \caption{}
        \label{fig:nadir}
    \end{subfigure}\hfill
    \begin{subfigure}{0.3\textwidth}
        \centering
        \includegraphics[width=\textwidth]{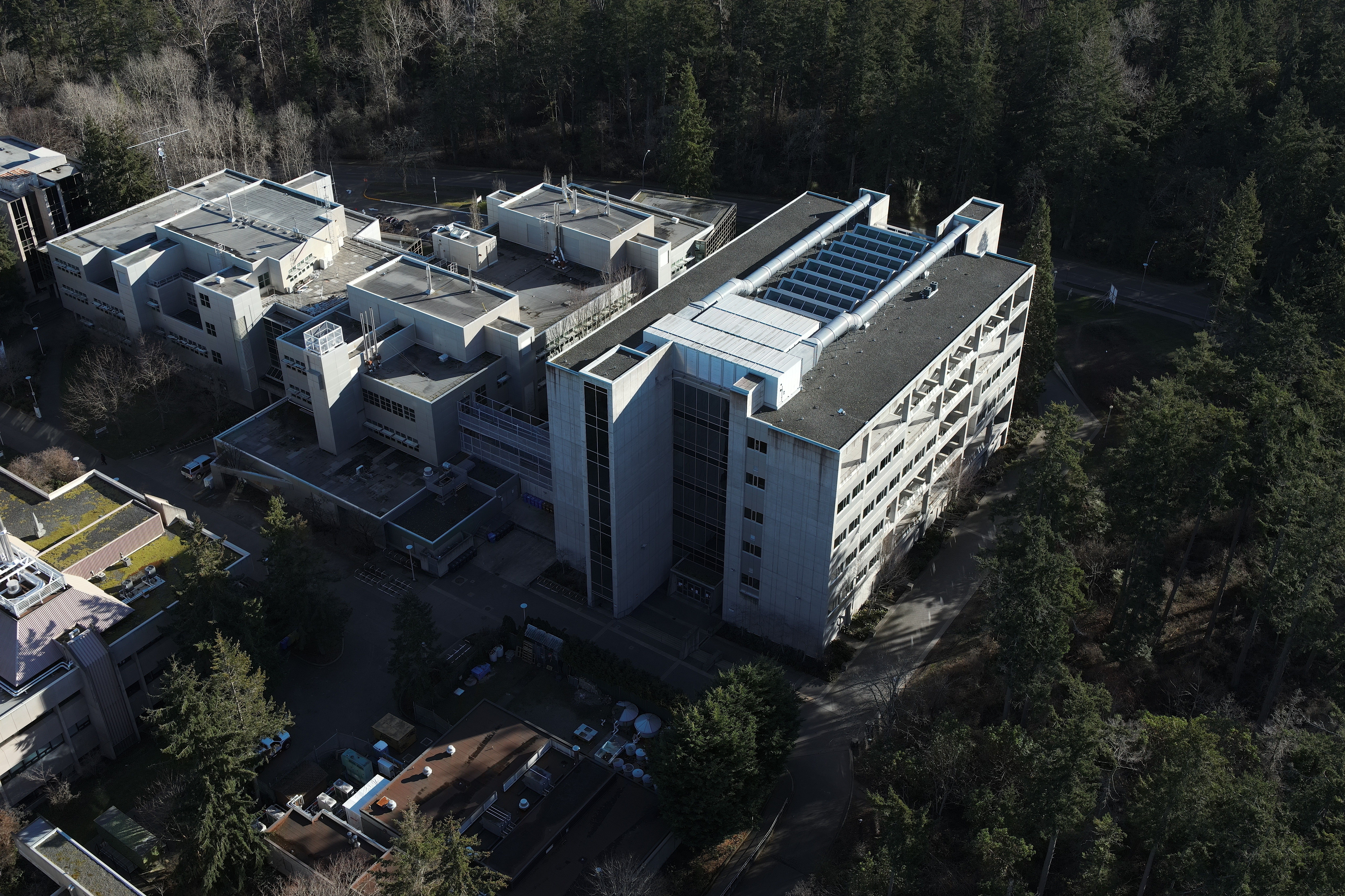}
        \caption{}
        \label{fig:oblique}
    \end{subfigure}\hfill
    \begin{subfigure}{0.3\textwidth}
        \centering
        \includegraphics[width=\textwidth]{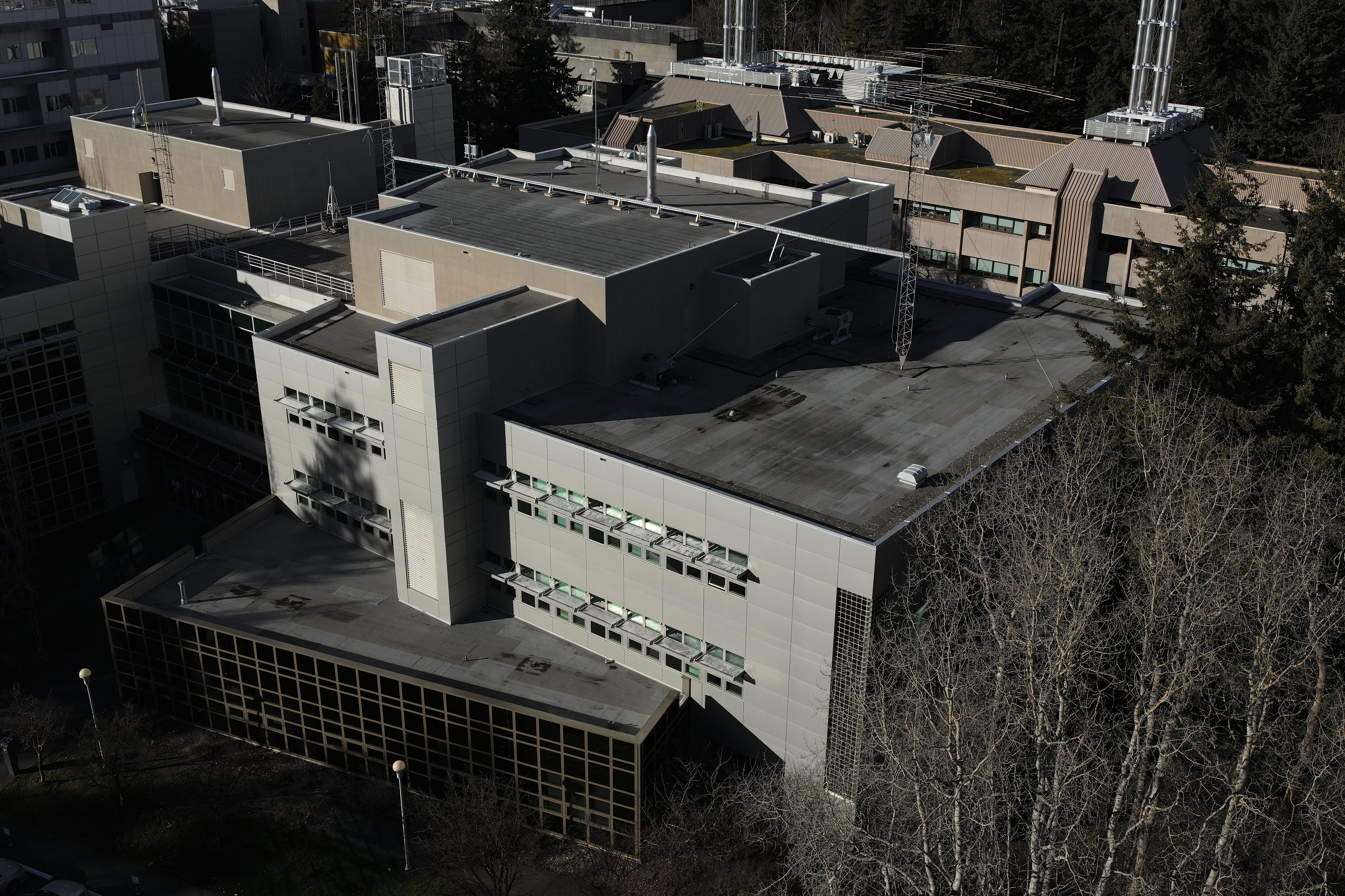}
        \caption{}
        \label{fig:closeup}
    \end{subfigure}

    \caption{Aerial images captured by the drone from (a) zenith, (b) oblique, and (c) close-up, separately.}
    \label{fig:aerialimages}
\end{figure}
In this section we describe our data acquisition and processing pipelines used to assess the suitability of using DRAC as a research platform for NeRF exploration and detailing our later move to Azure.

\subsection{Data Acquisition}
The aerial imagery focused on the Engineering Computer Science Building located at the University [name removed for blind review]. The equipment that we used to do the photogrammetry was a DJI M300 with a P1 full-framed survey camera, which uses real-time kinematics (RTK) to encode precise locations of the images during data collection. The flight paths are pre-planned parallel s-shaped paths that completely cover the entire building and surrounding area and are automatically executed by the flight control system. To acquire the best data quality, every two adjacent aerial images in the sequences had an 80\% overlap. 

The entire flight lasted about 40 minutes and resulted in high-quality imagery, however, sunshine conditions are present with some resulting shadows on the North side. In addition, Ground Control Points (GCPs) were measured by a professional surveyor and used in the processing pipelines described below. In total, we captured 844 high-resolution images including 426 images that are standard nadir scans done on the program, 397 oblique scans for 3D post-data processing structures, and 21 images taken manually to have a closer perspective of the building. Figure~\ref{fig:aerialimages} shows example images of the (a) zenith scan, (b) oblique scan, and (c) close-up, respectively\footnote{These images are made available to the research community at our share on [removed for blind review]}.

\begin{figure}[h]
    \centering
    \includegraphics[scale=0.165]{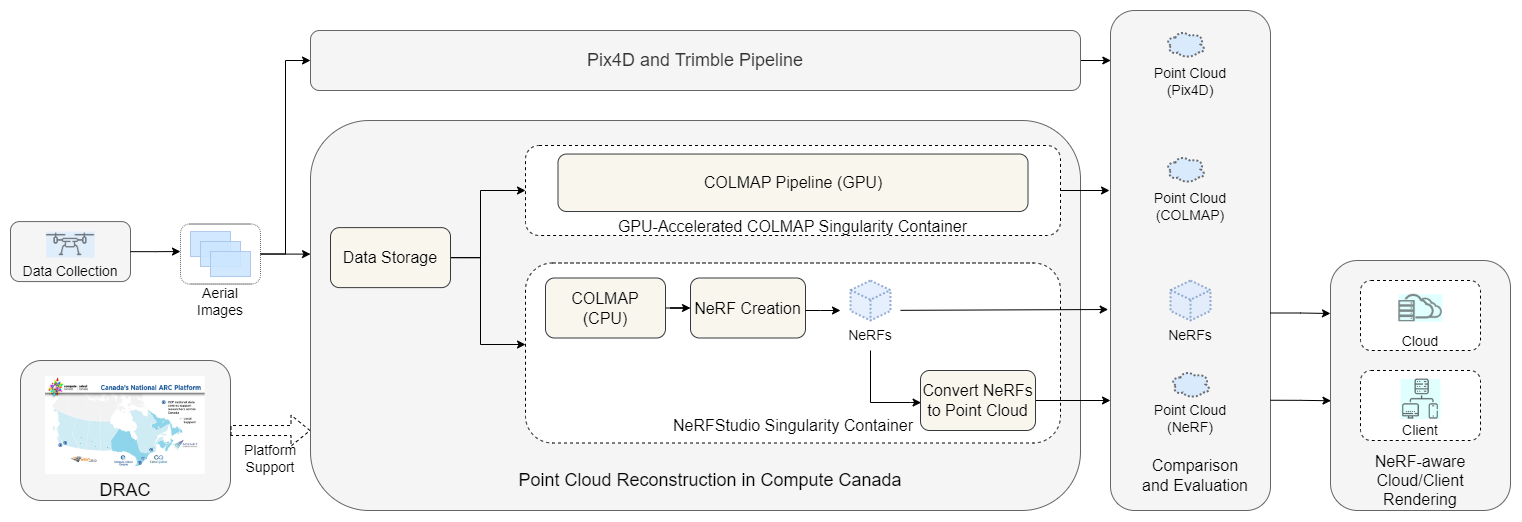}
    \caption{Pipelines for point cloud data reconstruction (from left to right), showing data acquisition through to three different point cloud representations and NeRFs. Specifically, Singularity containers on DRAC were used to deploy Nerfstudio (with COLMAP on CPU), alongside a GPU accelerated COLMAP pipeline.  Quality metrics were determined for the different point clouds.  The NeRFs were finally rendered on NeRF-aware platforms.}
    \label{fig:pipeline}
\end{figure}

\subsection{Processing Pipelines}
As illustrated in Figure~\ref{fig:pipeline}, the aerial images were processed by three different pipelines:  

{\em Pix4D and Trimble Business Center pipeline:} The drone operator and surveyor provided this pipeline of tools through which they normally deliver aerial surveys to their clients. The selection of Pix4D as the photogrammetry software used by the company that collected our professional data is not necessarily due to it having the best image stitching ability, but rather that it integrates an in-flight tool to automate the collection of the data. This allows the pilot to know when sufficient high quality data has been collected, and also allows a pre-filing of a flight plan for approval. This flight-time tool integrates well with a post-processing step in Pix4D to create a point cloud from the collected imagery. The surveyor then took the Pix4D data and ran an additional series of corrections on the data in Trimble Business Center software, which allows the resultant point cloud to include corrections from the ground control points. 

{\em COLMAP GPU accellerated pipeline:} COLMAP is open source software that performs the assembly of images into a point cloud. We use it as an alternative to Pix4D that generates a point cloud directly. The overlap between images is an important element to enable COLMAP to stitch together properly. Our base model of 100m nadir imagery (426 images) was used to then add oblique imagery at 100m, 70-80m, and 30m. 

{\em Nerfstudio pipeline:} We use Nerfstudio to train a custom model from the imagery. This software is a plug-and-play API that provides a user-friendly tool for creating, training and rendering NeRFs. In addition, it also provides a complete CLI for processing imagery data into the COLMAP structure for later use in model training and NeRF rendering.  Unlike the GPU accelerated pipeline using COLMAP described above however, this instance of COLMAP only uses CPU, and is part of the Neural Radiance Field (NeRF) construction process within the Nerfstudio package. 

\subsection{Azure Implementation}
This section describes the deployment of the pipeline on Microsoft Azure in two different formats:
\begin{enumerate}
    \item Azure Spot VM
    \item Azure Container Instances
\end{enumerate}
These two types of deployment model have different characteristics, hence valuable metrics and observations can be gathered. Azure Spot VM is quite cheap in terms of hourly cost, but it has a long cold-start time with the possibility to be evicted at any time on a 30-second notice. Azure Container Instances costs more, but the workload is guaranteed to finish. Being a VM itself, Azure Spot VM requires full manual operation while Azure Container Instances are cloud managed, thus saving human efforts.

Our implementation uses an end-to-end scheduling mechanism that operates as an intermediary layer connecting the data source and the ultimate NeRF model. In this approach, the imagery containing both RGB data and corresponding geographical or positional information is obtained from a continuously running server. This server exposes a series of endpoints enabling the submission of raw RGB data along with the positional input. Notably, the server actively monitors and manages available hardware resources, dynamically scheduling the model training workload according to demand. Leveraging the metrics and measurements obtained from various deployment models, the training component is delegated to different hardware resources based on performance estimations.

\subsubsection{Raw Data Storage}

Neural Radiance Fields (NeRFs) rely on either a collection of raw images or a single multi-angle video as their primary input. These input data typically exhibit a substantial size, often ranging from 1 to 2 gigabytes (GiBs). However, when utilizing the NeRF pipeline on Compute Canada, one encounters a constraint wherein the data storage is ephemeral and subject to routine cleanup by system administrators. Consequently, in order to ensure reliable creation and modification of the raw input dataset, it becomes imperative to establish a long-term storage solution to accommodate these data. Another challenge arises during the process of uploading the data from a client, as the inbound network capability of Compute Canada is limited due to security considerations. Therefore, the client needs to negotiate with a pre-existing server that facilitates hole-punching techniques to establish a connection with the processing server. To overcome this issue, we have opted to employ Azure Blob Storage as our designated raw data storage solution. This service offers convenient functionalities for uploading, storing, and mounting arbitrary binary data. The storage endpoint has been established on the cloud, with each processing node capable of mounting the storage on-demand for seamless access to the data.

\subsubsection{Processing and Training Nodes}

To execute the training and processing tasks, Azure Spot VMs have been chosen due to their optimal balance between high performance and cost-effectiveness. Recognizing the computationally intensive nature of the NeRF model, GPU-based machines equipped with NVIDIA T4 GPUs have been specifically allocated for this purpose, incorporating CUDA drivers and Docker container environments. These servers expose their inbound Docker TCP port to enable remote endpoint control and workload scheduling. Additionally, input and output storage are automatically mounted prior to each run and subsequently unmounted after completion to maintain consistent state across multiple runs. Resource preservation measures have also been implemented, with the scheduler server orchestrating automatic cold start and de-allocation of these machines.

\subsubsection{Orchestrator Node}

To tackle the challenges of accessibility and scalability, a centralized scheduler server has been implemented to manage all available resources and initiate them as needed. This scheduler server operates in conjunction with an orchestrator, which autonomously assigns NeRF processing or training workloads to computational nodes based on their current usage. In the event that no suitable node is available, the system dynamically allocates a new node to enhance the system's capabilities. The orchestrator assumes responsibility for overseeing the entire operation and maintaining metadata records to facilitate this process. Furthermore, the orchestrator node exposes multiple endpoints to enable client interaction, granting clients the capability to view or download the final results.

\subsubsection{Metadata Storage}

An inherent limitation of the NeRF pipeline on Compute Canada was the lack of a robust mechanism to track individual NeRF models throughout their lifecycle. As a consequence of the aforementioned accessibility constraints, it was not feasible to establish a metadata tracking system, which necessitated manual efforts for various NeRF creation processes. However, in Azure, we have employed a managed Azure PostgreSQL database to address this challenge by storing the highly relational data associated with NeRF models. The orchestrator node continuously polls and updates the database to ensure a consistent state, thereby maintaining an accurate and comprehensive record of metadata. This metadata encompasses essential information regarding each NeRF training request submitted to the system, as well as pertinent details related to their storage and processing.

\subsubsection{Workflow Description}
\begin{figure}[!t]
\centering
\includegraphics[width=\linewidth]{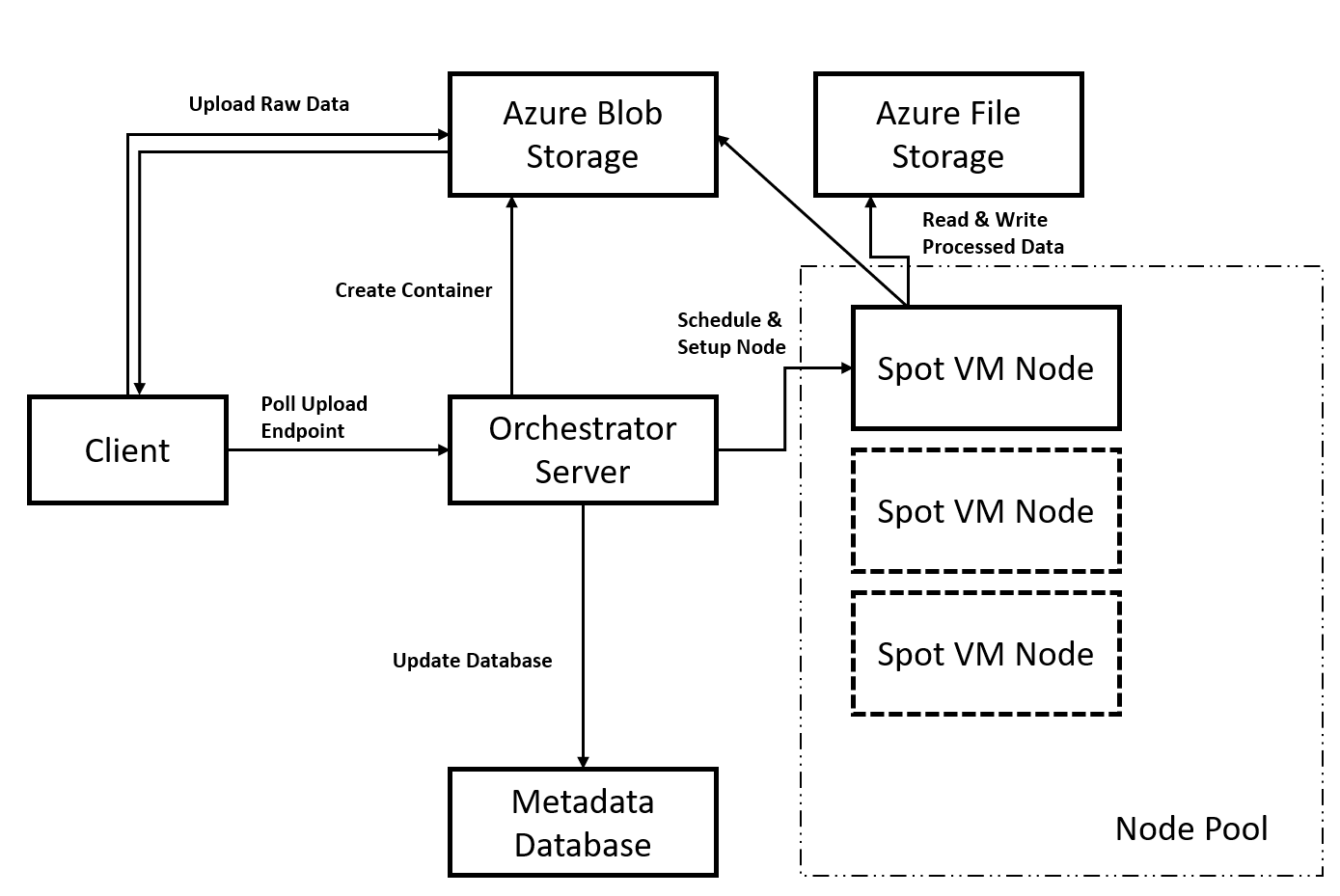}
\caption{Azure pipeline for NeRF processing, showing orchestrator, database, and storage in relation to compute nodes.}
\label{fig:azure_pipeline}
\end{figure}

As shown in Figure \ref{fig:azure_pipeline}, the client initiates the process by invoking an endpoint provided by the orchestrator server to obtain another endpoint for uploading their raw data. The orchestrator server allocates a blob storage container for the client and returns the corresponding upload endpoints. The client uses this endpoint to upload the raw data.

Once the upload is completed, the client informs the orchestrator server of the completion and provides descriptive information about the raw data. The orchestrator server updates the database with this information and sets up the storage accordingly.

The orchestrator server determines the availability of computational resources and either allocates a new Spot VM server or reuses an existing one. The input and output storage are then mounted on the compute node.

The orchestrator server spawns a processing container on the allocated server to handle the training workload.

After the training process is completed, the orchestrator server updates the database to maintain a consistent state and notifies the client of the task's completion. The data is stored long-term, allowing the client to update or view it as needed.

\section{Results}
\subsection{Resource Usage}

\begin{table}
\centering
\caption{Resource Usage}
\begin{tabular}{ |c|c|c| } 
 \hline
  & \textbf{Data Processing}  & \textbf{Model Training} \\ 
 \hline
 Peak CPU (\%) & 90 & 65 \\ 
 Avg CPU (\%) & 27 & 17 \\ 
 Peak Memory (MB) & 1507 & 2543 \\
 \hline
\end{tabular}

\label{table:Azure_metrics}
\end{table}

For the quantitative results, two test workloads were prepared and executed using the experiment pipelines. The first being a camera-recorded footage of a University campus located in California. The second one is the built-in poster NeRF dataset used to train the final model. All resources are allocated on demand with the orchestrator and metrics are collected on a per-machine basis. These results are shown in Table \ref{table:Azure_metrics}.

Average cold-start time for a spot machine is also measured from 10 individual restarts and shown in Table \ref{table:restart}.

\begin{table}[]
\centering
\caption{Cold Start Times}
\begin{tabular}{ |c|c| } 
 \hline
 \textbf{Test Num.}  & \textbf{Time (seconds)} \\ 
 \hline
 1 & 70.787  \\ 
 2 & 70.489 \\ 
 3 & 70.492  \\
 4 & 70.411 \\
 5 & 70.411\\
 6 & 70.495\\
 7 & 70.414 \\
 8 & 70.638 \\
 9 & 70.462 \\
 10 & 70.575 \\
 \hline
\end{tabular}

\label{table:restart}
\end{table}

\subsection{Scheduling And Orchestration}

The scheduling and orchestration process is facilitated through the computation of a penalty score, which incorporates various factors related to individual machines. These factors encompass the average CPU and memory utilization of the machines, coupled with a predetermined cold start penalty. Cost considerations also play a pivotal role in the calculation of the penalty score. Additionally, a user-adjustable bias parameter is incorporated within the formula, thereby allowing for meticulous fine-tuning based on specific cost-saving or batch-processing requirements. This comprehensive approach ensures the efficient allocation of computational resources while accommodating diverse operational priorities.

The penalty score is calculated as followed:
\begin{gather*}
    Penalty_{cpu} = 160^{\%_{avg}cpu} + |0.5 - \%_{avg}cpu * 80|^{\%_{avg}cpu} \\
    Penalty_{mem} = 160^{\%_{avg}mem} + |0.5 - \%_{avg}mem * 80|^{\%_{avg}mem} \\
    Penalty_{cold} = 100 \\
    Penalty_{cost} = 200 \\
    Penalty = (Penalty_{cpu} + Penalty_{mem} + \\ Penalty_{cold}) * bias + \\ Penalty_{cost} * (1 - bias)
\end{gather*}

The Azure pipeline is therefore set up to minimize processing costs while starting processing a set of client data within minutes rather than the hours that were taken for node allocation on the HPC cluster. The processing times are still long enough that it should by no means be considered a real-time system, or even one where the user will patiently wait for results while using the application, but that will return a NeRF from a set of input data in a reasonable time given processing cost constraints.

\subsection{Point Cloud Quality}
Assessing the quality of the NeRF directly is problematic, but we took the approach of instead analyzing the quality of a point cloud derived from the NeRF. A preliminary view of quality metrics was obtained by applying the approach proposed in \cite{zhang_no-reference_2022}.  The results of comparing the Nerf-derived point cloud to the Pix4D  and the Sensat Urban point clouds is presented in Table \ref{tab:point_cloud_comparison}. 

\begin{table}[h]
\centering
\caption{Comparison of point clouds for Pix4D, NERF, and SensatUrban \cite{hu_sensaturban_2022}}
\begin{tabular}{lccc}
\hline
\textbf{Cloud} & \textbf{Pix4D} & \textbf{NERF} & \textbf{SensatUrban} \\
\hline
Curvature  & 0.09 & 0.08 & 0.03 \\
           & (std dev: 0.06) & (std dev: 0.05) & (std dev: 0.02) \\
           & (entropy: 7.18) & (entropy: 7.09) & (entropy: 6.92) \\
\hline
Anisotropy & 0.84 & 0.86 & 0.96 \\
           & (std dev: 0.12) & (std dev: 0.11) & (std dev: 0.06) \\
           & (entropy: 6.77) & (entropy: 6.66) & (entropy: 5.27) \\
\hline
Linearity  & 0.48 & 0.51 & 0.45 \\
           & (std dev: 0.18) & (std dev: 0.19) & (std dev: 0.21) \\
           & (entropy: 7.31) & (entropy: 7.33) & (entropy: 7.41) \\
\hline
Planarity  & 0.35 & 0.35 & 0.51 \\
           & (std dev: 0.18) & (std dev: 0.18) & (std dev: 0.22) \\
           & (entropy: 7.27) & (entropy: 7.28) & (entropy: 7.44) \\
\hline 
Sphericity  & 0.16 & 0.14 & 0.04 \\
           & (std dev: 0.12) & (std dev: 0.11) & (std dev: 0.06) \\
           & (entropy: 6.77) & (entropy: 6.66) & (entropy: 5.31) \\
\hline 
{\bf Overall}  & {\bf 0.249} & {\bf 0.299} & {\bf 0.499} \\
          
\end{tabular}

\label{tab:point_cloud_comparison}
\end{table}

The overall result was calculated through regression of these features against the model in \cite{zhang_no-reference_2022}.  Though it is unsurprising that Sensat Urban has the highest quality of the three, these results show that the NeRF generated point cloud is of a competitive quality with the industry standard, Pix4D.

\section{Conclusion}
There were two primary aspects of this work that are worth noting in these conclusions. First, that the pipeline process on Compute Canada and Azure produced equivalent end results and the choice of platform rests primarily on operational requirements. The Azure platform allowed finer grained control over process execution and allowed better monitoring of the progress of the pipeline. Second, the overall quality of point cloud creation through an intermediary NeRF represenatation was of at least as good a quality as the direct creation of the point cloud within the Pix4D software. The use of reference-free metrics and differences between the data sets does not support as direct a comparison to the SensatUrban-derived point clouds except to note that the data quality of the SensatUrban point clouds were an explicit goal of their collection process, which was more rigorous than that used in this study. \cite{hu_sensaturban_2022}
 
\section{Discussion and future work}

Although this paper has focused on the presentations of two pipelines for processing NeRF data from the client, most of our future work entails what it is what we do with this pipeline. The constraints of the DRAC HPC cluster really precluded our using it for an on-demand processing pipeline from the client. The more flexible storage and compute orchestration available on Azure allows us to begin experimenting with those various client applications. The capture of scenes in their entirety is of great interest in our virtual production work. In many cases we will use drone-based data acquisition methods and capture a full large-scale scene all in one dataset, but that works best when the eventual use of the scene is for aerial shots. The less data there is in the original dataset from a particular view, the noisier and more hallucination-prone is the resultant NeRF. Which means that for many storytelling purposes ground views of a scene are preferred. We are currently experimenting with an iPhone-based client that uses the lidar on the iPhone Pro to provide a camera position so that the COLMAP step of the above pipeline can be avoided, thus reducing the computational cost of training the NeRF. We are also working on decomposing the scene NeRF into a series of object NeRFs that we can positionally rebuild into a scene. This allows us to provide a pre-computed starting point for NeRF generation rather than using a flat prior for each new scene.
As these efficiencies accumulate, we get closer to the vision of an iPhone based client that can send imagery and positional data to the cloud, and in return quickly get a multi-view representation of the scene that has been captured. This is relevant both to formal filmmaking activities like virtual production, but also to more app-centric uses like background replacement and XR device displays.
These applications are particularly attractive in establishing a sense of place for digital storytelling efforts. As we get the client application working, we are also in the midst of building a studio-based virtual production environment in place where actors can come and place themselves realistically in scenes that others record. This is one step closer to the vision of being able to re-create arbitrary aspects of the real world and mix them with both human and animated characters without the careful setup constraints of traditional greenscreen compositing.

\section*{Acknowledgment}

Thank you to the National Research Council of Canada for funding this study, and to the Digital Research Alliance of Canada for providing compute resources.



\bibliography{IEEEabrv,spring.bib,summer.bib}

\end{document}